\pdfoutput=1

\documentclass{ecai} 

\usepackage{latexsym}
\usepackage{amssymb}
\usepackage{amsmath}
\usepackage{amsthm}
\usepackage{booktabs}
\usepackage{enumitem}
\usepackage{graphicx}
\usepackage{color}

\newcommand{\BibTeX}{B\kern-.05em{\sc i\kern-.025em b}\kern-.08em\TeX}

\begin{document}


\begin{frontmatter}


\paperid{123} 


\title{Parallel Strategies for Best-First Generalized Planning}


\author[A]{\fnms{Alejandro}~\snm{Fernández-Alburquerque}\orcid{0009-0009-0884-7015}}
\author[A]{\fnms{Javier}~\snm{Segovia-Aguas}\orcid{0000-0001-6141-2672}} 

\address[A]{Universitat Pompeu Fabra}


\begin{abstract}
In recent years, there has been renewed interest in closing the performance gap between state-of-the-art planning solvers and generalized planning (GP), a research area of AI that studies the automated synthesis of algorithmic-like solutions capable of solving multiple classical planning instances. One of the current advancements has been the introduction of Best-First Generalized Planning (BFGP), a GP algorithm based on a novel solution space that can be explored with heuristic search, one of the foundations of modern planners. This paper evaluates the application of parallel search techniques to BFGP, another critical component in closing the performance gap. We first discuss why BFGP is well suited for parallelization and some of its differentiating characteristics from classical planners. Then, we propose two simple shared-memory parallel strategies with good scaling with the number of cores.
\end{abstract}

\end{frontmatter}


\section{Introduction}

\textit{Generalized planning} (GP) has been a longstanding area of research in Artificial Intelligence (AI) \citep{jimenez-2019, srivastava-2011b}. The foundation of GP is \textit{automated planning}, which studies how to construct sequences of actions (commonly known as \textit{plans}) to go from a specific initial state to a goal \citep{ghallab-2016}. Since planning is a hard problem (PSPACE-complete) \citep{baz-2021}, solving multiple problem instances from the same domain is computationally expensive. Given this realization, GP studies the synthesis of general plans that can solve multiple problem instances from the same domain, reducing the computational complexity to a one-time up-front cost \citep{hu2011generalized, srivastava-2011}.

The state-of-the-art planning solvers are often heuristic-based planners \citep{taitler20242023}. These planners guide the combinatorial search of reaching a goal state from the initial state with heuristics, usually based on computing the cost of solving a \textit{relaxed plan} as an estimate of the actual cost of reaching the solution \citep{BONET20015}. Given the success of heuristic search in planning, \citet{segovia-aguas-2021,segovia-aguas-2024} propose a heuristic-based approach to generalized planning, which they call Best-First Generalized Planning (BFGP). BFGP leverages a set of novel GP native heuristics and a new solution space, independent of the number of input instances, to compute general algorithmic solutions.

Parallel programming is tightly coupled with AI's recent success, as CPU manufacturers have transitioned to multi-core processors due to single-core performance stagnation \citep{kishimoto-2009}. As a consequence, there has been a notable effort to parallelize fundamental algorithms like Best-First Search (BFS), and these techniques have already been successfully applied to planning domains \citep{baz-2021, kishimoto-2009, kuroiwa-2021}. One of the most impactful algorithms has been Hash Distributed A* (HDA*) \citep{kishimoto-2009}, which efficiently handles the two most challenging communication overheads of Parallel BFS: distributing search nodes and keeping a consistent closed list of reached nodes. To achieve this, they use a hash function to assign each state to a unique process (which allows local duplicate detection) and asynchronous communication.

While using parallel techniques for classical planners is an active research topic, applications to generalized planning are much less explored. In this paper, we discuss how the BFGP algorithm can be easily parallelized by design. Furthermore, we present two shared-memory parallelization strategies that can scale linearly with the number of cores.


\section{Suitness of Best-First Generalized Planning for parallelization}

A \textit{classical planning} problem \citep{haslum2019introduction} is defined as  $P = \langle{\cal D}, {\cal I}\rangle$, where ${\cal D} = \langle F,A\rangle$ is the domain that comprises the set of lifted predicates $F$ and actions $A$, and ${\cal I} = \langle \Omega, I, G\rangle$ is the instance that specifies the set of constant objects, the initial state $I$, and the goal condition. A solution to $P$ is a sequence of actions or plan $\pi$ that maps the initial state to a goal state where the goal condition holds. 

GP is formalized as the problem of finding an algorithm-like solution $\Pi$ (also known as generalized plan) to a set of $T$ planning problems, that is ${\cal P} = \{P_1,P_2,\ldots,P_T\}$, where all planning problems share the domain, but may differ in the instances (different objects, initial state and/or goal condition). We focus on a special kind of generalized plans named planning programs. Formally, a planning program \citep{segovia-aguas-2016} is a sequence of $n$ instructions $\Pi = \langle w_0, ..., w_{n-1}\rangle$, where each instruction $w_i$ is associated with a \textit{program line} $0 \leq i < n$ and can be of one of the following three types:

\begin{description}
    \item[A planning action] $w_i \in A$, where $A$ is the set of deterministic functions of the planning domain.
    \item[A goto instruction] $w_i = goto(i', !y)$, where $i'$ is a program line such that $0 \leq i' < n$ and $i' \ne i$, and $y$ is a proposition. Proposition $y$ can be the result of an arbitrary expression on state variables.
    \item[A termination instruction] $w_i = \texttt{end}$. The last instruction of a planning program is always a termination instruction.
\end{description}

A planning program $\Pi$ is a solution to ${\cal P}$ iff the execution of $\Pi$ on every $P_i\in{\cal P}$ generates a classical plan $\pi_i$ that is a solution to the original planning problem. 

In this work, we use the generalized planner BFGP \citep{segovia-aguas-2024}, which has shown good performance for computing planning programs via heuristic search. BFGP is a \emph{frontier-search} algorithm \citep{korf2005frontier}, which means that it does not repeat states during the search (i.e., all expanded planning programs are different). This unique characteristic eliminates the need to maintain a closed list of states to prevent search loops. Notably, this simplifies the algorithm's parallelization, as duplicate detection is a significant source of synchronization and communication overheads in parallel BFS.

Another relevant detail about BFGP is that it is a \emph{Greedy Best-First Search} (GBFS). Unlike A*, the GBFS is guided by an evaluation function $f(n) = h(n)$ that only takes into account the estimated cost of reaching the goal from node the current node $n$ (instead of also considering the solution cost up to the current node, $f(n) = g(n) + h(n)$) \citep{kuroiwa-2021}. In a GBFS, we are only interested in finding a solution to the problem, not necessarily the optimal one. Likewise, this trait eases the task of parallelizing BFGP since once we find a solution, it is unnecessary to check for optimality. 


\section{Parallel Best-First Generalized Planning}

The following section presents our two strategies for parallelizing BFGP\footnote{All experiments used BFGP settings in \citep{segovia-aguas-2022} and were run on a 12-core Intel® Core™ i7-12700 Processor (20 threads) and 32G of RAM.} and evaluates their performance in $9$ different classical planning domains; $3$ of them are propositional (\textit{corridor}, \textit{gripper}, and \textit{visitall}) and the other $6$ are numeric domains (\textit{fibonacci}, \textit{find}, \textit{reverse}, \textit{select}, \textit{sorting}, and \textit{triangular sum}). 

\begin{table}[ht]
\caption{Execution time (seconds) of the first parallel strategy for different numbers of threads ($t$). $t=1$ is the original BFGP. Best results in bold.}
\vspace{5mm}
\centering
\begin{tabular}{ll@{\hspace{8mm}}ll} 
\toprule
\textbf{Domain} & \multicolumn{3}{c}{Execution time} \\
\midrule
& $t=1$ & $t=2$ & $t=4$ \\
\cmidrule(lr){2-4}
Corridor & 1212.37 & 619.8 & 296.2 \\
Fibonacci & 0.46 & 0.24 & 0.13 \\
Find & \textbf{0.0} & \textbf{0.0} & \textbf{0.0} \\
Gripper & 12.92 & 6.72 & 3.16 \\
Reverse & 0.08 & 0.04 & 0.02 \\
Select & 0.02 & 0.02 & 0.02 \\
Sorting & \textbf{0.01} & \textbf{0.01} & \textbf{0.01} \\
Triangular Sum & \textbf{0.0} & \textbf{0.0} & \textbf{0.0} \\
Visitall & 338.14 & 227.85 & 157.91 \\\\

& $t=8$ & $t=16$ \\
\cmidrule(lr){2-3}
Corridor & 15.24 & \textbf{12.37} \\
Fibonacci & 0.08 & \textbf{0.03} \\
Find & \textbf{0.0} & \textbf{0.0} \\
Gripper & 1.28 & \textbf{0.59} \\
Reverse & \textbf{0.01} & \textbf{0.01} \\
Select & \textbf{0.01} & \textbf{0.01} \\
Sorting & \textbf{0.01} & \textbf{0.01} \\
Triangular Sum & \textbf{0.0} & \textbf{0.0} \\
Visitall & 114.91 & \textbf{79.48} \\

\bottomrule
\label{tab:results_strat_1}
\end{tabular}
\end{table}

\textbf{Parallel strategy \#1.} It sequentially expands nodes until there are at least $N$ nodes per thread\footnote{$N$ is a parameter of the algorithm.}. Then, it starts a parallel search in which each thread is independent and does not share nodes with other threads. To ensure a balanced workload distribution, $N$ should be larger for planning domains that require more program lines to reach a solution. Table~\ref{tab:results_strat_1} evaluates the scaling of this strategy (the single-threaded execution corresponds to the original BFGP implementation).

\begin{table}[ht]
\caption{Execution time (seconds) of the second parallel strategy for different numbers of threads ($t$). Best results in bold.}
\vspace{5mm}
\centering
\begin{tabular}{ll@{\hspace{8mm}}ll} 
\toprule
\textbf{Domain} & \multicolumn{3}{c}{Execution time} \\
\midrule
& $t=4$ & $t=8$ & $t=16$ \\
\cmidrule(lr){2-4}
Corridor & 768.89 & 442.1 & \textbf{40.81} \\
Fibonacci & 0.13 & 0.1 & \textbf{0.05} \\
Find & \textbf{0.0} & \textbf{0.0} & \textbf{0.0} \\
Gripper & 5.7 & \textbf{2.88} & 3.96 \\
Reverse & 0.04 & \textbf{0.02} & \textbf{0.02} \\
Select & 0.02 & 0.02 & \textbf{0.01} \\
Sorting & \textbf{0.01} & \textbf{0.01} & \textbf{0.01} \\
Triangular Sum & \textbf{0.0} & \textbf{0.0} & \textbf{0.0} \\
Visitall & 118.82 & 113.16 & \textbf{20.84} \\

\bottomrule
\label{tab:results_strat_2}
\end{tabular}
\end{table}

\textbf{Parallel strategy \#2.} In this strategy, threads distribute promising nodes during the parallel search phase, so there is a tradeoff between searching the most promising states and minimizing the communication overhead. In our solution, we compute the cost-to-go value of each generated node, and if it is equal to or better than the last expanded node, we send the new node to another thread. In contrast to HDA*, where a hash function determines which process will receive the node, we simply cycle between all threads. This approach is viable because BFGP does not need to perform duplicate detection. Table~\ref{tab:results_strat_2} evaluates the scaling of this strategy\footnote{The single-threaded results of Table~\ref{tab:results_strat_1} also apply to Table~\ref{tab:results_strat_2}.}.





\vspace{-0.6em}

\section{Discussion}

The first strategy performs very well, with speedups ranging from $\sim$4x to $\sim$98x in the most complex domains. Furthermore, increasing the number of threads always results in better performance. On the other hand, no parallel strategy strictly dominates. In some domains (like \textit{Visitall}), the second strategy gets better scaling and performance than the first strategy, but in others, it gets slower execution times. We believe that a better prioritization of promising nodes and the use of asynchronous communication would help the second strategy perform better than the first one. To conclude, our results show that BFGP is well-suited for parallelization, and further developments could make BFGP capable of handling more complex problems from IPC planning domains \cite{taitler20242023}.






\vspace{-0.6em}

\bibliography{main}

\end{document}